\documentclass[conference]{IEEEtran}
\IEEEoverridecommandlockouts
\usepackage{cite}
\usepackage{amsmath,amssymb,amsfonts}
\usepackage{algorithmic}
\usepackage{graphicx}
\usepackage{textcomp}
\usepackage{url}
\usepackage{comment}
\usepackage{xcolor}
\def\BibTeX{{\rm B\kern-.05em{\sc i\kern-.025em b}\kern-.08em
    T\kern-.1667em\lower.7ex\hbox{E}\kern-.125emX}}
\begin{document}

\title{
Topic Modelling of Swedish Newspaper Articles about Coronavirus: a Case Study using Latent Dirichlet Allocation Method \\
\thanks{This work has been partially supported by grant EP/V047949/1 "Integrating hospital outpatient letters into the healthcare data space" (funder: UKRI/EPSRC).
$^*$ \textit{co-first authors}
}
}

\author{\IEEEauthorblockN{Bernadeta Griciūtė$^*$ }
\IEEEauthorblockA{\textit{Saarland University} \\
\textit{University of Malta }\\
Germany \& Malta \\
griciute.bernadeta@gmail.com}
\and
\IEEEauthorblockN{Lifeng Han$^*$}
\IEEEauthorblockA{\textit{Department of Computer Science} \\
\textit{University of Manchester}\\
Manchester, UK \\
lifeng.han@manchester.ac.uk}
\and
\IEEEauthorblockN{Goran Nenadic}
\IEEEauthorblockA{\textit{Department of Computer Science} \\
\textit{University of Manchester}\\
Manchester, UK \\
g.nenadic@manchester.ac.uk}
} 

\maketitle

\begin{abstract}
Topic Modelling (TM) is a natural language processing (NLP) method for discovering topics in a collection of documents. Being an unsupervised method, it is a valuable tool when trying to summarise the main topics and topic changes in large quantities of data.
In this study, we apply two prevalent topic modelling techniques - Latent Dirichlet Allocation (LDA) and BERTopic - to analyse the change of topics in the Swedish newspaper articles about COVID-19.
We describe the corpus we created including 6515 articles, methods applied, and statistics on topic changes over approximately 1 year and two months period of time from 17th January 2020 to 13th March 2021.
We hope this work can be an asset for grounding applications of topic modelling and can be inspiring for similar case studies in an era with pandemics, to support socio-economic impact research as well as clinical and healthcare analytics. 
\textit{Our data and source code is openly available at \url{https://github.com/aaronlifenghan/Swed_Covid_TM}.}
\end{abstract}


\section{Introduction}

During the Coronavirus (COVID-19) pandemic \footnote{World Health Organisation (WHO) COVID-19 Dashboard \url{https://covid19.who.int/} 6,656,601 deaths and 651,918,402 confirmed cases till 23 December 2022.}, when the majority of the countries imposed strict lockdowns, the Swedish government instead adopted a relatively different and even controversial approach towards Coronavirus compared to other countries especially during the ``first wave'' such as keeping many sectors open in society \cite{creutz2021nordic_pandemic_travel,hedman2022dying_covid_sweden,kubai2022sweden_covid_image}. \footnote{\url{https://www.government.se/search/?query=covid} 
} 
There have been some studies on the impact of the Swedish COVID-19 policy such as the effect on overall infection rates, death rates, and the vulnerability towards older people \cite{pashakhanlou2022sweden_corona_public}, on the criticism and argument about lack of scientific guidance from the government \cite{brusselaers2022evaluation_sweden_covid}, on the potential influence towards other kinds of disease on nation and region levels \cite{saarentausta2022potential_covid_sweden}, and on the spread level among different personnel, e.g. dental sector \cite{fredriksson2022prevalence_covid_dental_sw}. 

Researchers from the statistical background also carried out  mathematical modellings on the prediction of spatiotemporal risks towards incidence, intense care (IC) admission, and death, e.g. Bayesian analysis by \cite{jaya2022joint_Bayesian_sweden}.

However, there has not been enough comprehensive research work on the socio-economic impact of computational social science fields. In addition, for future healthcare text analytics studies, there are yet many investigations to be carried out on debates and discussions from the society around this topic and policy. 
In this work, to further facilitate this research direction, we leverage the methodologies from natural language processing (NLP) and 
carry out an experimental investigation on topic modelling using Swedish Newspaper articles about Coronavirus. This aims at getting more insight into the topic focuses and changes around Coronavirus in Swedish society over more than a year  from 17th January 2020 to 13th March 2021. 

The method we applied to this study is an unsupervised statistical generative model called Latent Dirichlet Allocation (LDA), which was first designed by \cite{blei2003latent_LDA} to address text modelling, classification, and collaborative filtering tasks. LDA method has been proven to be very effective in topic discovery and similar text identification, in the topic modelling field since then \cite{rehurek_lrec_Gensim_topic2010,tong2016text_LDA_TM,asmussen2019smart_review_TM}.  

The rest of the paper is organised as below: Section \ref{sec_related} surveys the related work to ours on topic modelling in NLP, healthcare text analytics, and social media mining, Section \ref{sec_method} presents LDA method, Section \ref{sec_data_exp} follows up with our experimental work carried out on Coronavirus topic using LDA, and Section \ref{sec_discussion} concludes the paper with discussion and future work.

\section{Related Work}
\label{sec_related}
We present related work on topic modelling (TM) methods briefly in NLP, Healthcare Text Analytics (HTA), and TM in Journalism and Social media research.


TM research started in the early 2000s as a breakthrough in machine learning (ML) techniques to address the automatic analysis of large amounts of digital (electronic) archived documents.
These methods used hierarchical probabilistic models including LDA by \cite{blei2003latent_LDA}, Markov Topic Models (MTM) by \cite{wang2009markov_TM}, and their variations such as including authorship features by \cite{rosen2004author_TM},
Dynamic Topic Models (DTM) by \cite{blei2006dynamic_TM,Blei_2012ProbTM}, which incorporated the evolution feature of the topics based on LDA often using a corpus with a sequence of time stamps, for instance, using journal articles. 

Despite the popularity of LDA, \cite{gerlach2018network_TM} designed a new approach to carry out TM by looking into community networks using a ``stochastic block model'' (SBM). The SBM was designed to automatically detect the ``number of topics'', which parameter has to be manually set up in the LDA algorithms. 
With the new development of NLP methodologies, especially the continuous vector space representation of lexical semantics \cite{google2017attention,devlin-etal-2019-bert}, in very recent years, researchers also tried to approach the TM task in this track. Representative work includes BERT-Topic \cite{grootendorst2022bertopic_tf_idf} and Top2Vec \cite{angelov2020top2vec}. TM in continuous vector space is still an emerging direction.


In Healthcare Text Analytics (HTA), 
\cite{kovavcevic2012topic_suicide_rule_ML} applied both rules and machine learning based methods to investigate topic categorisation of statements from suicide notes. 
\cite{SPASIC2014605_2014cancer_mining} carried out survey work on cancer research models applied to text mining from NLP. 
\cite{noble2021using_topic_dogs} used electronic health records (EHRs) to identify the outbreak of dog disease in the United Kingdom. Some recent advances in HTA using clinical discharged summary letters are reported by \cite{Han_Wu_etal2022_PLM4clinical} from data-constrained fine-tuning comparing different pre-trained language models.

In journalism, Newspaper, and Social media mining field, \cite{jacobi2016quantitative_journalistic_TM} applied LDA algorithms to the New York Times articles covering nuclear technology from 1945 to 2010s for topic trend and pattern analysis.

There are also research projects that made efforts on creating such datasets and shared tasks to facilitate the advancement of TM in social-media healthcare. For instance, the ``Social Media Mining for Health (SMM4H)-2017 shared task'' \cite{sarker2018data_twitter_task} organisers prepared 15,717 annotated tweets for classifying adverse drug reactions, and 10,260 tweets for classifying medication consumption. 
Some earlier work on social media web crawling for terminological and lexical research, and topic modelling can be found at \cite{greenwood2008lexical_web_topic,greenwood2009prioritising_hyperlink_term}.

Recently, \cite{piksina2020coronavirus_stock_swedish} carried out an experimental investigation on the Swedish stock market change due to Coronavirus using social media data and the LDA method via the ``Konstanz information miner'' Analytics Platform \cite{10.1145/1656274.1656280_KNIME_miner}. 
However, to the best of our knowledge, there isn't published research work on a comprehensive investigation and analysis of Swedish COVID19 policy on socio-economical impact using LDA methods. In our investigations, we will report the topic modelling results on several different categories including public opinions on the origins of the virus, governmental health recommendation policy, scientific research sector, and economy.

\section{Revisiting LDA Method}
\label{sec_method}

The concept of generative model LDA can be described as below \cite{blei2003latent_LDA,Blei_2012ProbTM}:

\begin{gather*} 
p(\beta_{1:K},\theta_{1:D},z_{1:D},w_{1:D})\\
=\Pi_{i=1}^Kp(\beta_i)\Pi_{d=1}^Dp(\theta_d) \\
\left(\Pi_{n=1}^Np(z_{d,n}|\theta_d) p(w_{d,n}|\beta_{1:K},z_{d,n})\right)
\end{gather*}

\noindent where the four main parameters $\beta$, $\theta$, $z$, and $w$ represent respectively the ``topic distribution'', ``topic proportion of document'', ``topic assignment of document'', and the ``observed words of document''.
For instance, 1) $\beta_k$ from $\beta_{1:K}$ can be a vocabulary distribution of words, i.e., their statistical probabilities; 
2) $\theta_d$ from $\theta_{1:D}$ is the topic proportion of the \textit{d}th document, e.g. $\theta_{d,k}$ can be the ``topic proportion'' of $\beta_k$ reflected in the \textit{d}th document;
3) $z_d$ from $z_{1:D}$ is the ``topic assignment'' to the \textit{d}th document, e.g. $z_{d,n}$ can be the ``topic assignment'' of the \textit{n}th word in the \textit{d}th document; and
4) $w_d$ from $w_{1:D}$ is the set of observed words in the \textit{d}th document, e.g. $w_{d,n}$ can be the \textit{n}th word observed from the \textit{d}th document.

The calculation is based on conditional probability theories, i.e. the ``topic assignment'' event $z_{d,n}$ is conditioned on the ``topic proportion'' $\theta_d$, and the ``word observation'' $w_{d,n}$ is conditioned on all the topics $\beta_{1:K}$ and the topic assignment $z_{d,n}$.

The ideal computation is to sum all the joint distribution across all the possibilities among the potential topics. However, this computation is so huge that alternative solutions are often used to approximate it. There are two popular methods including sampling and variation \cite{Blei_2012ProbTM}.
The difference is that sampling-based methods try to approximate it using collected samples and forming an empirical distribution \cite{gladkoff-etal-2022-measuring}, while variation methods aim at structuring the task as an optimisation challenge \cite{blei2003latent_LDA}.   

\section{Topic Modelling on Coronavirus}
\label{sec_data_exp}

\subsection{Corpus Preparation}


Since the first wave of COVID-19 in Europe, the term ``Coronavirus'' has stayed in the main headlines of the media. How has the focus in Coronavirus related articles shifted with time regarding different stages of the lockdown and the exponentially rising infection numbers?
The Swedish government's decision on non-harsh lockdown but rather appearing to the common sense of people seeking herd immunity made it stand out from the nearby countries. But how has this decision been depicted and commented on in local media?
To investigate these questions, we chose to create a corpus from articles published by Sveriges Television (SVT) \footnote{\url{https://www.svt.se}}, Sweden's national public television broadcaster funded by a public service tax. The SVT is not the biggest news site in Sweden and is less popular than commercial ``Expressen'' or ``Aftonbladet''. However, we expect it to be more neutral and have fewer click-bite articles since it is funded by the tax.

We created our corpus by scraping an SVT webpage in which Coronavirus related articles were collected \footnote{\url{https://www.svt.se/nyheter/utrikes/25393539}}. As of 13th of March 2021 when this research project was done, there were 6,515 Covid19 related articles with the first one coming from the 17th of January 2020, which covered more than one year period.

To visualise the distribution of these articles per time, we created a graph depicting the number of articles published each day related to COVID from 2020/01/17 to 2021/03/31, which is 422 days overall in Figure \ref{fig:numb_article_time}.
As shown in the figure, the number of published articles peaked, in the beginning, two months of the year, then steadily decreased towards the summer of 2020 and rose again in autumn when the second wave of the virus arrived. This visualisation can be further improved in future work by investigating how many percentages of all published articles including COVID and non-COVID ones had the Corona term constituted.\textit{}

\subsection{Text Pre-Processing}
For our experiments, we have chosen a period of exactly one year, dating from the publication of the first COVID article 2020/01/17 to 2021/01/17. 
To ensure a higher percentage of the articles including Coronavirus as the topic and the diversity of content, we filtered out local news (``lokalt'') since they have many repeated content, and only kept nationwide (``inrikes'') and foreign (``utrikes'') categories. 
The corpus consisted of 2,251 articles after filtering. \textit{} 

\begin{figure}[!t]
\centering
\includegraphics*[width=0.45\textwidth]{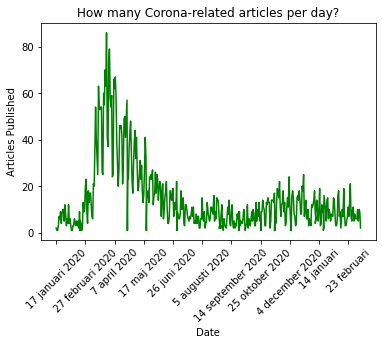}
\caption{Number of Published Articles with Timeline}
\label{fig:numb_article_time}
\end{figure}

For experimental purposes, we divided the corpus into 12 time frames counting from the 17th day of each month. The number of articles per time frame, following the tendency seen in the graph (Figure \ref{fig:numb_article_time}), was very uneven. The largest number of articles among these time-frames was from 17th March 2020 to 17th April 2020 including 569 articles, while the smallest number came from 17th August to 17th September with 23 articles only. 
While we initially planned to take the same number of articles for every month as an alternative solution, sorting in accordance with the real trend made it more objective.

As mentioned by \cite{asmussen2019smart_review_TM}, the LDA method requires several parameter setups including text pre-processing, selection of the number of topics to generate, and expert analysis of outputs, etc.

Text pre-processing was first carried out for topic modelling purposes using the Gensim toolkit by \cite{rehurek_lrec_Gensim_topic2010}, such as removing ``stop words'' and very rare words which might appear only once. 
Gensim used a vectorised dictionary including bigrams as part of this processing. \footnote{\url{https://radimrehurek.com/gensim/intro.html}} 

To perform tokenisation and summarise the words, we tried SnowBall Stemmer \footnote{\url{https://snowballstem.org/}} but the output was not ideal to our task. Words kept their suffixes, e.g. ``virus-et'' ( meaning ``the virus''), or were difficult to guess. The spaCy library \footnote{\url{https://spacy.io/}} also has references to a Swedish lemmatiser, but none of the implementations seemed to be available for usage. 
Finally, we lemmatised the corpus using Stanza lemmatiser \cite{qi-etal-2020-stanza_tool}
developed by the Stanford NLP group. \footnote{\url{https://stanfordnlp.github.io/stanza/}}


\subsection{More Parameters: number of topics}
Using LDA algorithms, the execution time of Gensim is relatively short for the smaller number of iterations. We have run it several times with different parameter values on the ``number of topics'', trying to inspect what could be the optimal case, which is not a straightforward task.

\begin{figure}[!h]
\centering
\includegraphics*[width=0.3\textwidth]{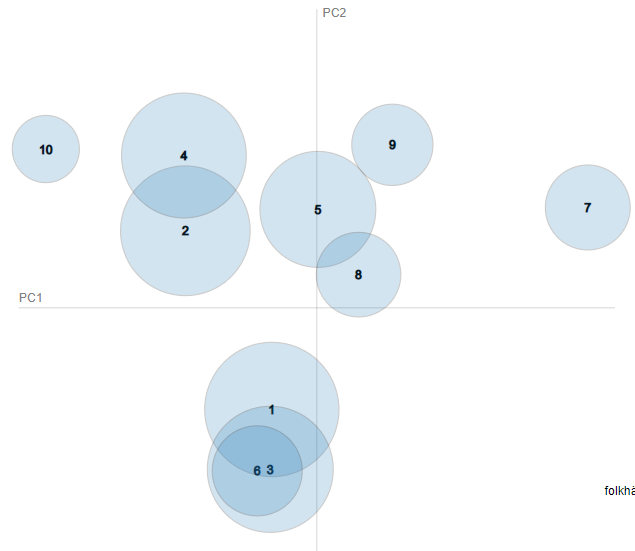} 
\includegraphics*[width=0.3\textwidth]{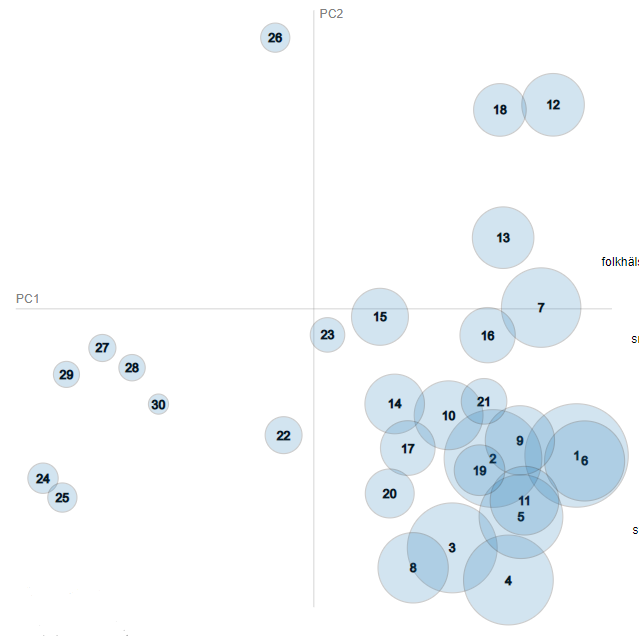}
\caption{Distribution Representation of 10 (up) vs 30 (down) Topics}
\label{fig:topic_10vs30}
\end{figure}

While topic coherence tests can be helpful to compute the optimal topic number, we have chosen a more ``human-in-the-loop'' method to ensure better quality output. We check the result representation ourselves manually by using different parameters, to see how overlapping the output topics can be. We tried various options between 10 to 50 topics, and it seemed that 15 to 25 topics are not too overlapping and meanwhile keep a good balance between being too general and too specific.
The visualisation of topic distribution for 10 and 30 topics is displayed in Figure \ref{fig:topic_10vs30}.

To carry out dynamic topic modelling (DTM), i.e. the change of topics over time, Gensim packages have two different implementations available, its own ``ldaseqmodel'' and a python wrapper for ``dtmmodel''. \footnote{available at \url{https://radimrehurek.com/gensim/models/ldaseqmodel.html} and \url{https://radimrehurek.com/gensim/models/wrappers/dtmmodel.html}} 
We have tried both of them and the biggest difference we observed seemed to be the running time. The ``ldaseq'' model runs for almost 8 hours while the ``dtm'' model only needs 12 minutes. This time difference is not fully comparable though, since the first model we ran was using un-lemmatised corpus which might result in more tokens to process.
We tried the ``ldaseq'' model with 15 topics and the ``dtm'' model with 15, 20, and 25 topics. The output of topics seemed quite similar between these models.
For visualisation, we used the pyLDA package plus the ``dtmvisual'' toolkit by \cite{brousseaudynamic_archives2019}. \footnote{\url{https://github.com/GSukr/dtmvisual}} 


\subsection{Outputs of LDA using Gensim}
Using LDA algorithms integrated with Gensim generally gave us well-grouped topics that have been summarised on what the country was breathing in during the examined time period. 
Among these topics, there are word groups related to the most vulnerable part of society, e.g. the ``old people'', ``infections and deaths'' in old people's homes. Other topics include ``children and school'' and ``home-teaching'', etc. The most prevalent topics include ``Health Ministry'', ``state epidemiologist A. Tegnell'', and ``death per day''.
There are also some economy-related topics, as well as talking about the ``origin of the virus'', or the situation in other countries.

\subsection{Classified Outputs using DTM}
\begin{figure*}[!h]
\centering
\includegraphics*[width=0.95\textwidth]{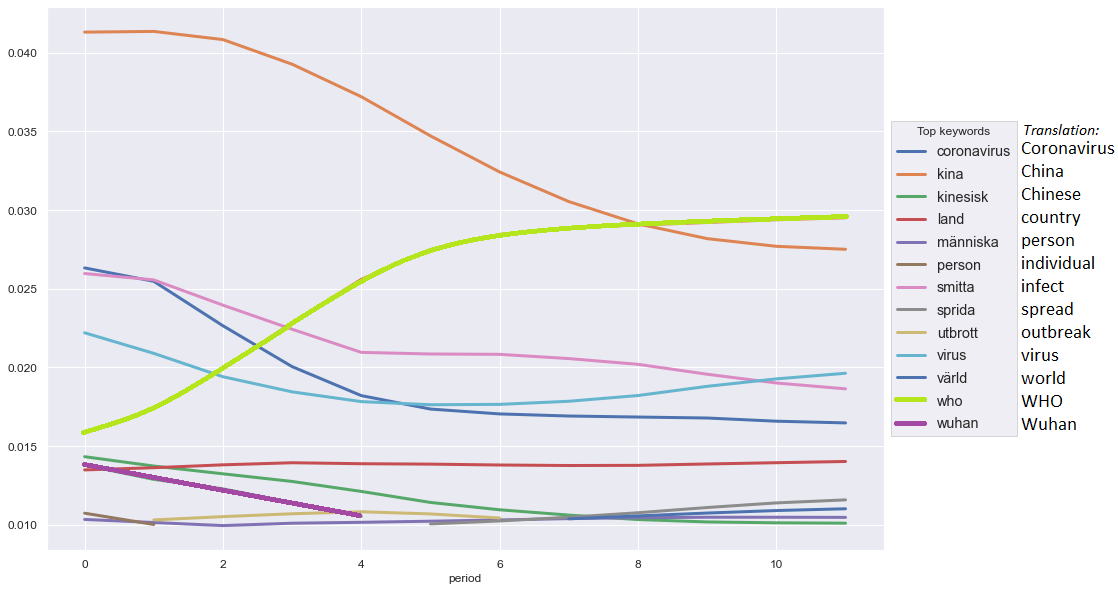}
\caption{Topics on China (kina) and WHO with Timeline}
\label{fig:kina_WHO}
\end{figure*}

\begin{figure*}[!h]
\centering
\includegraphics*[width=0.95\textwidth]{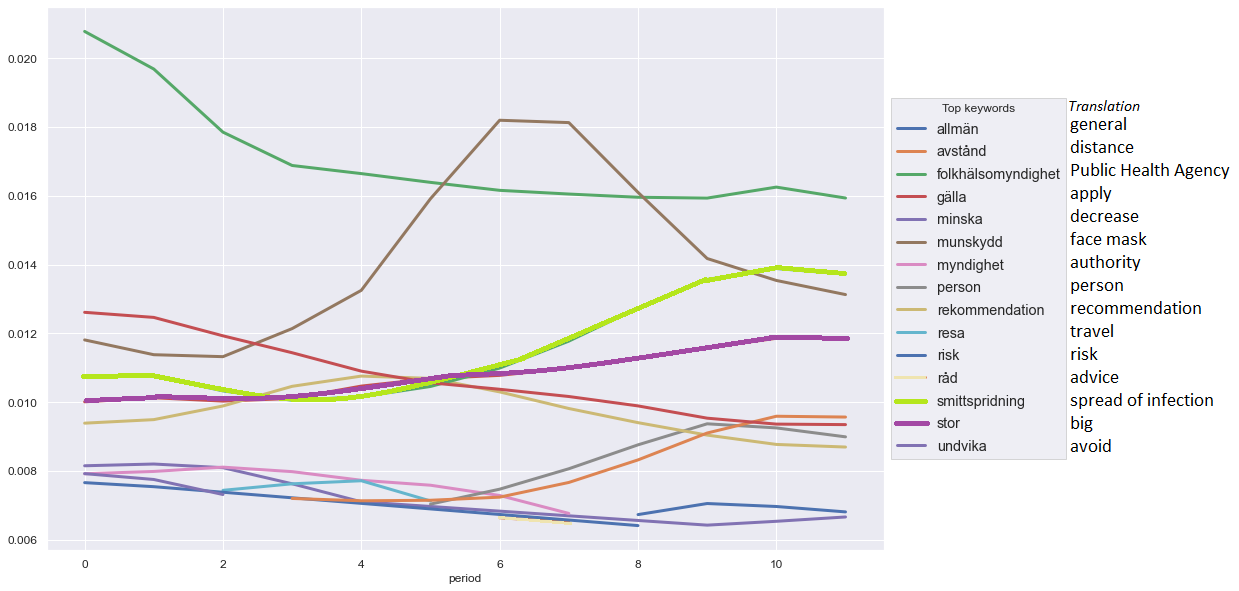}
\caption{Topics Related to Public Recommendations with Timeline}
\label{fig:mask_spread}
\end{figure*}

\begin{figure*}[!h]
\centering
\includegraphics*[width=0.95\textwidth]{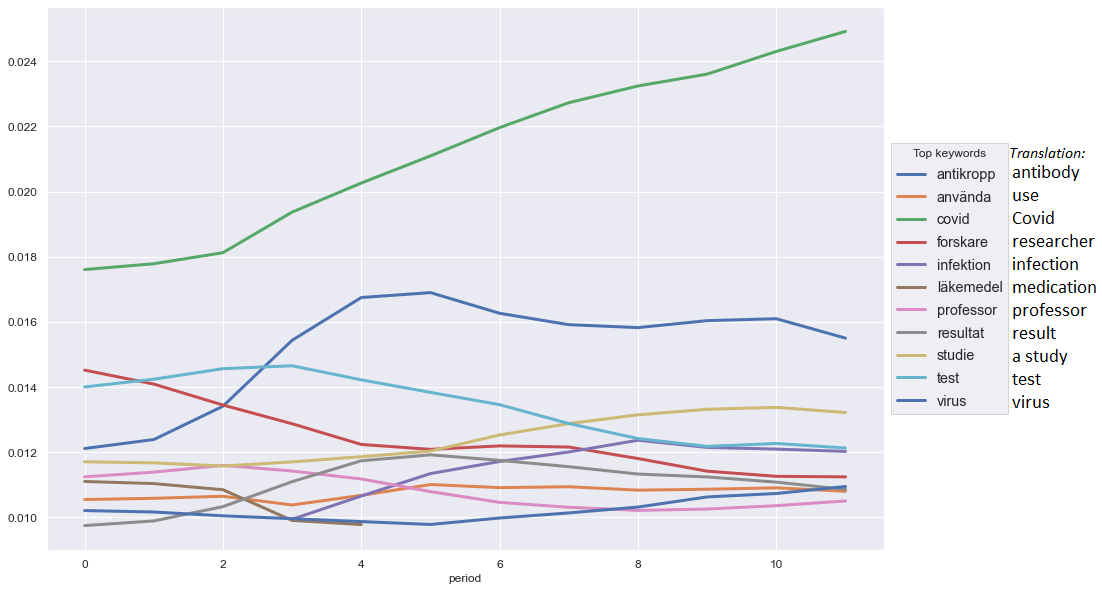}
\caption{Topics Related to Antibody Research with Timeline}
\label{fig:antibody_research}
\end{figure*}

\begin{figure*}[!h]
\centering
\includegraphics*[width=0.95\textwidth]{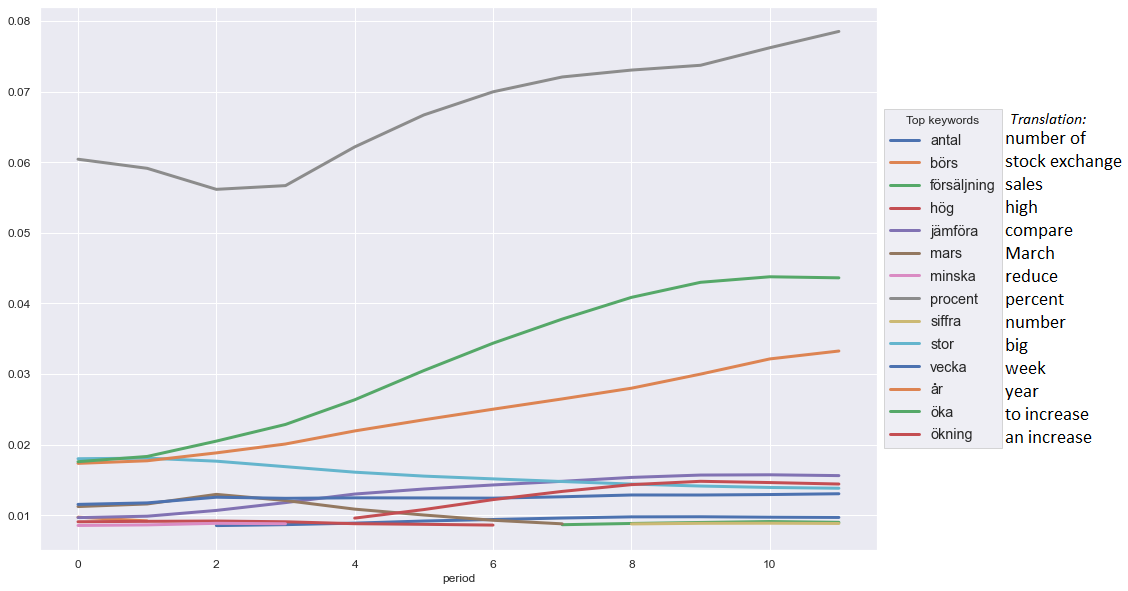}
\caption{Topics Related to Economy with Timeline}
\label{fig:economy_sale}
\end{figure*}

We list some example outputs using dynamic topic modelling (DTM) below with 20 trained topics from the following categories: 1) the origin of the virus and WHO strategies, 2) public health recommendations, 3) antibody scientific research, and 4) economy. In Figure \ref{fig:kina_WHO} to \ref{fig:economy_sale}, the more frequent keywords are ranked on the top levels, so as to the color lines in the figures. For instance, if there are two lines with the same color, the upper-level line corresponds to the upper-level keyword in the keyword list. 

In Figure \ref{fig:kina_WHO}, the graph shows how Coronavirus related topics shifted from talking about China (``kina'') and ``Wuhan'' to the World Health Organisation (WHO). This reflected people's attention from discussing the virus's origin to how WHO was addressing the issue with what strategies.

Figure \ref{fig:mask_spread} demonstrates the topics on public recommendations, where indicated a drastic change in the frequency of face masks (``munskydd'') which reflected the change in people's attitude towards this suggestion. 
There is a decrease in the mention of the Public Health Agency (``folkhälsomyndighet'') but an increase in spread and infection (``smittspridning''), which is coherent with people's concerns more about the social impact of the virus. 
We also observed a decrease in ``recommendation'' (\textit{rekommendation}) while an increase in ``advice'' (\textit{råd}).

Figure \ref{fig:antibody_research} shows a different perspective from  ``researcher'' (\textit{forskare}),  ``antibody'' (\textit{antikropp}), and ``drugs'' (\textit{läkemedel}) related discussion. 
The blue curve on ``antibodies'' showed a sharp increase in the beginning months of the studying period, then stayed as an influential topic. This reflected either the trust or debates in society regarding biomedical scientific research. 

From a socio-economic perspective, Figure \ref{fig:economy_sale} shows how the concerns on the economy increased over time, including the keywords ``stock exchange'' (\textit{börs}) and ``sales'' (\textit{försäljing}), both of which curves grew steadily.

\section{Discussion and Future Work}
\label{sec_discussion}


In this work, to understand more about the societal impact of the Swedish government on their policies towards Coronavirus, we carried out an experimental investigation using topic modelling (TM) methodology on Swedish newspaper articles covering around one year period. 
We first introduced the pandemic background of our work and related scientific research on this topic. 
Then we explained the corpus we collected and the mathematical models of latent Dirichlet allocation (LDA) we applied for this study including the toolkits we used. 
We finally presented the topic modelling outputs using LDA and dynamic topic modelling (DTM) by classifying them into several catergories from the topics on the origin of the virus and WHO strategies, to public health recommendations, scientific biomedical research, to the economic discussion. 
The outcomes proved the successful applications of LDA method in our research task.

In future work, there are many aspects that can be carried out to directly improve this study. 
Firstly, it is worthy to collect articles published by different newspaper agents and forums. 
For instance, some newspaper agents might have more to say about the opinions from pop stars on the lockdown, e.g. the ``Aftonbladet.se'', while others might have more radical criticism on the governmental strategies.

Secondly, data collected from different countries can also enrich the experimental outputs, especially from nearby Scandinavian countries. For instance, Denmark and Norway both chose a stricter lockdown during the pandemic, shutting down the borders with Sweden and criticising the tactic of the latter. Very often, Scandinavian countries talk about each other in their news articles, sharing some related words in the topics under concern, which gives a chance to build a comparable corpus. Furthermore, beyond Scandinavian countries, we also have interests in collecting data from English and Chinese speaker countries, e.g UK and China, where there were also very strict lockdowns implemented by the central governments. We expect to extract more diverse \textit{health-related} topics in these countries that had strict lockdowns. 
For instance, in England,
a)  teenagers were kept home instead of going to school and playgrounds during Covid19 and this has been reported to cause them much stress and depression.
b) Teenagers and kids have to face some domestic violence during the lockdown, especially in some families where their parents used to fight when kids were away, but during the lockdown, it was unavoidable. This leaves post-Covid traumas and longer influence on the kids' mental health.
c) Many people died during the Covid19 and their relatives and family were not allowed to attend the funerals, which not only left long-term pain but also regret that ``they were not there'' during the funeral of loved ones.
d) Kids were separated from grandparents due to strict policy, but this affects the boundness between kids and grandparents.
Instead, e) in China, when the harsh lockdown suddenly arrived, many parents who worked in the city could not relocate back to their hometowns in villages, which left their kids separated from their parents and got more bound to grandparents who were living with them in the villages.

Thirdly, in the technical components, some further steps can be done to optimise the model, e.g. extending the stop word list and removing the least and most common words, optimising the number of topics via the criterion of topic coherence, and investigating the performance of different models available on LDA, e.g. Mallet toolkit \footnote{\url{https://rare-technologies.com/tutorial-on-mallet-in-python/}} and ``infinite DTM''. 

Extending from this case study, there are also many research directions to be explored. 
Firstly, ambiguous word analysis and cross-lingual TM \cite{boyd2010linguistic_TM} can be carried out to enhance model performance by borrowing corresponding techniques from NLP fields.

Secondly, ``multi-word token (MWT)'' expansion features in Stanza tool \cite{qi-etal-2020-stanza_tool} can be further investigated by using advanced studies on  multi-word and implicit expressions including their connection to emotionality \cite{han-etal-2020-alphamwe,han-etal-2021-chinese,han-2022-phd-thesis,griciute-etal-2022-cusp,piccirilli-schulte-im-walde-2022-features}.

Thirdly, as a departure from statistical probabilistic models, we plan to explore the word embedding space and neural models. For instance, the ``Word2Vec'' and ``FastText'' options in Gensim packages can be a bridge from word vectors to neural structures such as BERT-topic \footnote{our initial output using BERT-topic is displayed in Appendix}. 


Fourthly, it will be interesting to carry out experimental investigation on ``exchangeable topic modelling'' in a multi-nominal situation \cite{blei2006dynamic_TM}, and see how different topics interact with each other.

Finally, from the evaluation perspective, qualitative evaluation of the LDA and topic modelling research was always demanded in the field \cite{nikolenko2017topic_qualitative,Blei_2012ProbTM}, 
e.g. there are spaces for expert-based human-in-the-loop evaluations \cite{han2022overview_mte,han_gladkoff_metaeval_tutorial2022,gladkoff-han-2022-HOPE} to be carried out on the coherence levels within extracted topics. 
How to better evaluate model confidence levels \cite{gladkoff-etal-2022-measuring}, and the interpretation of the algorithms are other research directions.

\section*{Acknowledgment}

The author BG thanks Prof. Dr. Alexander Koller for advice on the experimental work.
We thank Christine from Manchester Council for the valuable feedback towards our future work discussion on children's mental health from Covid19 influence. We thank Hao Li for the BERT-topic model output, and thank anonymous reviewers for valuable comments for improving our paper.

\vspace{12pt}

\section*{Appendix}
\begin{figure*}[!h]
\centering
\includegraphics*[width=0.95\textwidth]{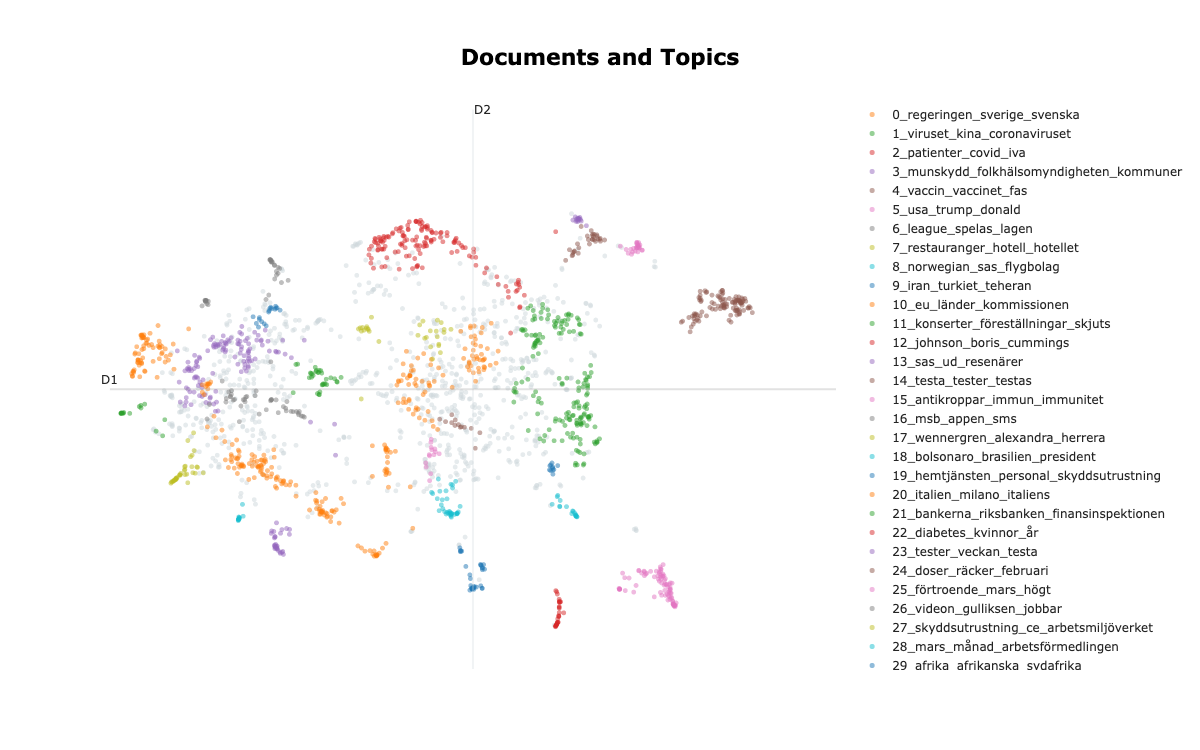}
\caption{BERT-Topic Output of Clustering using Minimum 10 Sentences per Cluster}
\label{fig:Bert-10sent}
\end{figure*}

\begin{figure*}[!h]
\centering
\includegraphics*[width=0.95\textwidth]{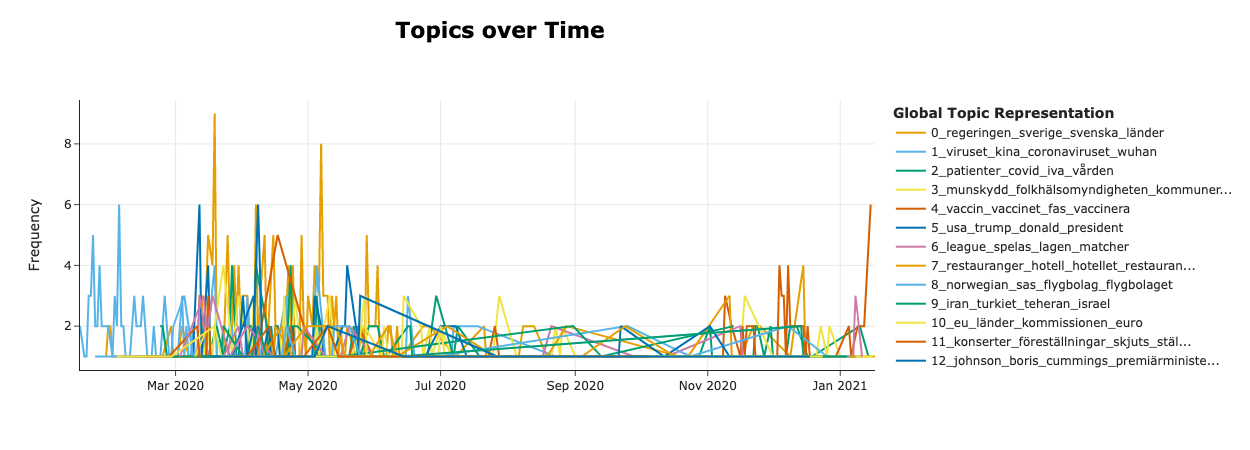}
\caption{BERT-Topic Output of Clustering using Minimum 10 Sentences per Cluster with Time Frame}
\label{fig:Bert-10sent-time}
\end{figure*}

\begin{figure*}[!h]
\centering
\includegraphics*[width=0.76\textwidth]{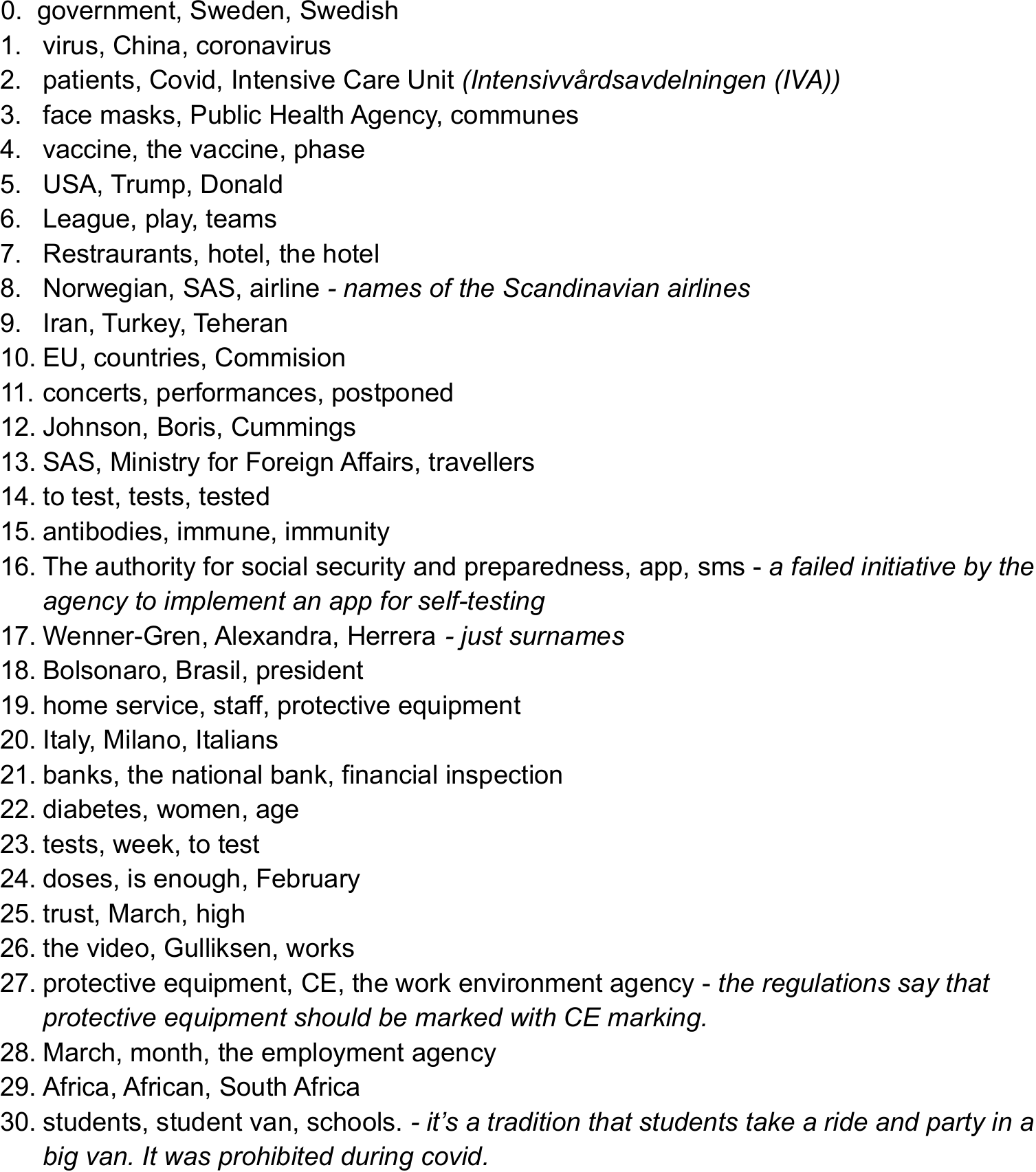}
\caption{English Translation of the 30 Classes from  Minimum 10 Sentences per Cluster from Figure \ref{fig:Bert-10sent}}
\label{fig:BERTtopic-10sent-translate}
\end{figure*}

We present our initial clustering outputs using BERT-Topic \cite{grootendorst2022bertopic_tf_idf} 
models on our dataset with a minimum of 10 sentences per cluster in Figure \ref{fig:Bert-10sent} and \ref{fig:Bert-10sent-time}, and translation in Figure \ref{fig:BERTtopic-10sent-translate}.

\end{document}